\documentclass[11pt,a4paper]{article}
\usepackage[table]{xcolor} 
\usepackage{graphicx}
\usepackage[hyperref]{emnlp2020}
\usepackage{times}
\usepackage{latexsym}
\usepackage{amsmath}
\usepackage{lipsum}
\usepackage{mathtools}
\usepackage{cuted}
\usepackage{booktabs}
\usepackage{array}
\usepackage[most]{tcolorbox} 
\usepackage{enumitem}
\usepackage{listings}

\usepackage{pifont}

\definecolor{lightgray}{gray}{0.95}
\definecolor{usercolor}{rgb}{0.95, 0.95, 1.0} 
\definecolor{botcolor}{rgb}{0.95, 1.0, 0.95} 

\newtcolorbox{userbox}{
  colback=lightgray, 
  coltext=black, 
  rounded corners,
  boxrule=0pt,
  boxsep=5pt,
  left=5pt,
  right=5pt,
  top=4pt,
  bottom=4pt,
  arc=10pt,
  auto outer arc,
  width=0.8\columnwidth, 
  halign=left, 
  enlarge left by=0mm, 
  enlarge right by=0mm
}

\newtcolorbox{botbox}{
  colback={rgb:red,33;green,138;blue,255}, 
  coltext=white, 
  rounded corners,
  boxrule=0pt,
  boxsep=5pt,
  left=5pt,
  right=5pt,
  top=4pt,
  bottom=4pt,
  arc=10pt,
  auto outer arc,
  width=0.8\columnwidth, 
  halign=left, 
  enlarge left by=10mm, 
  enlarge right by=-10mm
}

\newtcolorbox{botsuggestionbox}{
  colback=white, 
  coltext=gray, 
  rounded corners,
  boxrule=0pt,
  boxsep=5pt,
  borderline={0.5mm}{0mm}{blue!70!white,dashed},
  left=5pt,
  right=5pt,
  top=4pt,
  bottom=4pt,
  arc=10pt,
  auto outer arc,
  width=0.8\columnwidth, 
  halign=left, 
  enlarge left by=10mm, 
  enlarge right by=-10mm
}

\newtcolorbox{commandbox}{
  colback=white, 
  colframe=gray!30, 
  boxrule=1pt,
  boxsep=5pt,
  left=5pt,
  right=5pt,
  top=4pt,
  bottom=4pt,
  arc=5pt,
  auto outer arc,
  width=0.8\columnwidth, 
  halign=left, 
  before=\vspace{-15pt}, 
  after=\vspace{10pt}  
}

\usepackage{microtype}

\aclfinalcopy  

\makeatletter
\newcommand\footnoteref[1]{\protected@xdef\@thefnmark{\ref{#1}}\@footnotemark}
\makeatother
\newcommand{\printfnsymbol}[1]{%
  \textsuperscript{\@fnsymbol{#1}}%
}
\title{Task-Oriented Dialogue with In-Context Learning}
\author{Tom Bocklisch \quad Thomas Werkmeister \quad Daksh Varshneya \quad
Alan Nichol\thanks{\hspace*{0.5em}alan@rasa.com} 
\\ Rasa}

\date{}

\begin{document}
\maketitle
\begin{abstract}
We describe a system for building task-oriented dialogue systems combining the in-context learning abilities of large language models (LLMs) with the deterministic execution of business logic. LLMs are used to translate between the surface form of the conversation and a domain-specific language (DSL) which is used to progress the business logic. We compare our approach to the intent-based NLU approach predominantly used in industry today. Our experiments show that developing chatbots with our system requires significantly less effort than established approaches, that these chatbots can successfully navigate complex dialogues which are extremely challenging for NLU-based systems, and that our system has desirable properties for scaling task-oriented dialogue systems to a large number of tasks. We make our implementation available for use and further study\footnote{The approach described in this paper is implemented in Rasa under the name CALM: https://rasa.com/docs/rasa-pro/calm/}. 
\end{abstract}

\section{Introduction}
\label{sec:introduction}

The workhorse of industrial task-oriented dialogue systems and assistants is a modular architecture comprising three components: natural language understanding (NLU), dialogue management (DM)
\footnote{While in the research literature, the dialogue manager is often separated into dialogue state tracking (DST) and dialogue policy components, this distinction is rarely made in industrial applications.}
, and natural language generation (NLG) \cite{youngpomdp, young2007cued}.

Utterances spoken or written by end users are translated into \emph{dialogue acts}, where a dialogue act comprises an \emph{intent} and a set of \emph{entities}. 
For example, an utterance such as ``I need a taxi to the station" might be assigned to the intent \texttt{book\_taxi} and the entity \texttt{destination} with value ``station".
This dialogue act representation acts as the interface between the NLU and DM components of the system. 
The dialogue manager contains the logic to react to a \texttt{book\_taxi} intent by initiating a taxi booking task, prompting the end user for the time, pick-up location, etc.
These fields are typically called \emph{slots}.
As the dialogue progresses, subsequent user messages are also represented as dialogue acts, such as \texttt{inform(time=3pm)}.
The dialogue manager reacts to this sequence of inputs by executing actions and responding to the 
end user, either via a rule-based or a model-based dialogue \emph{policy}.
We refer to this as the intent-based NLU approach, and it is used by the major industry platforms for building chat- and voice-based dialogue systems like Rasa \cite{bocklisch2017rasa}, Dialogflow\footnote{https://cloud.google.com/dialogflow}, Microsoft Luis\footnote{https://www.luis.ai/}, and IBM Watson\footnote{https://www.ibm.com/docs/en/cloud-private/3.1.0?topic=services-watson-assistant}.

\subsection{Limitations of an intent-based NLU approach}

A key feature of the intent-based NLU approach is that it poses natural language understanding as a classification task: messages are ``understood" by assigning them to a predefined intent.
This is a powerful simplifying assumption.
In theory, intents provide an interface that fully abstracts the language understanding component from the dialogue manager.

However, working with a fixed list of intents has limitations which become more pronounced as an application matures and scales:
\begin{itemize}
    \item The taxonomy of intents becomes difficult to remember and reason about when the number of intents reaches several hundred, complicating annotation \& feedback loops as well as application debugging.
    \item Because the dialogue manager is coded to expect specific sequences of intents, making changes to intent definitions and introducing new intents becomes increasingly error-prone, as shifts in classifier outputs introduce regressions.
    \item Intents are typically defined to map closely to the tasks the assistant can perform, but user utterances often do not correspond directly to a specific task. A developer may create intents like \texttt{replace\_card} and \texttt{block\_card}, but end users often describe situations in their own terms (e.g. ``I lost my wallet"), which could map to a number of different tasks. 
    \item Messages are assigned to the same intent irrespective of context.\footnote{It is possible to overcome this limitation either by adding heuristics to the NLU module, or by training a supervised model which includes conversation history in the input. In practice neither of these approaches has been widely adopted.}
\end{itemize}

Furthermore, salient information is often discarded when translating a surface form into a dialogue act, and it falls on the dialogue manager to  \emph{reinterpret} the output of the NLU module to account for context. For example, the dialogue manager may prompt the user with a yes/no question in order to fill a boolean slot (e.g. ``Would you like to proceed?"). When the end user's response is mapped to an \texttt{affirm} or \texttt{deny} intent, the dialogue manager reinterprets this output and sets the boolean slot to \texttt{true} or \texttt{false} respectively. Similarly, the NLU module might detect a generic entity like \texttt{person} or \texttt{location}, which is then mapped to a task-specific slot like \texttt{transfer\_recipient} or \texttt{taxi\_destination}.

\subsection{Desiderata}

Given the impressive capabilities of recent LLMs on language understanding benchmarks \cite{huang2022large} and in-context learning \cite{brown2020language}, we investigate whether a superior approach to task-oriented dialogue can be developed by reconsidering the split of responsibilities between the  NLU, DM, and NLG components. 

In this work we aim to develop a system for building industrial task-oriented dialogue systems with the following attributes:
\vspace{4pt}

\noindent \textbf{Fast iteration:} The system should allow for rapid prototyping and testing. The delay between making a change (e.g. modifying task logic) and testing it should ideally be measured in seconds.
\vspace{4pt}

\noindent \textbf{Short development time:} The system should provide general conversational capabilities out of the box, so that developers can focus on implementing their unique business logic.  
\vspace{4pt}

\noindent \textbf{Concise representation of business logic:}
Both developers and subject-matter experts with less technical knowledge should be able to create and modify task logic easily.
\vspace{4pt}

\noindent \textbf{Reliable execution of business logic:}
Business logic of arbitrary complexity should be executed reliably, i.e. we should not rely on a language model to remember and follow a set of steps and branching conditions. 
\vspace{4pt}

\noindent \textbf{Explainable and debuggable:}
It should be possible to explain why the system responded in a certain way at any given time. 
\vspace{4pt}

\noindent \textbf{Scalable to a large number of tasks:}
It is common for AI assistants in industry to support hundreds of tasks. The system needs to be able to identify the correct task out of hundreds of possibilities.
Maintaining the system and adding new tasks should not become more complex as the system increases in size. 
\vspace{4pt}

\noindent \textbf{Model agnostic:}
Progress in LLMs is rapid and the approach should allow developers to adopt the latest models without having to re-implement their business logic. 

\vspace{4pt}

\section{Related Work}
\label{sec:related_work}

Here we briefly describe some recent streams of research which have developed alternatives to the intent-based NLU paradigm for building task-oriented dialogue.

\subsection{End-to-End Learning}
The advent of seq2seq models \cite{sutskever2014sequence} gave rise to a number of end-to-end approaches, notably \cite{bordes2016learning} who used a synthetic task to study whether an end-to-end model could learn the business logic for a task-oriented dialogue system purely from conversation transcripts grounded in knowledge base queries. 
In this stream of work, models are learned end-to-end, but assistant responses selected from a candidate list rather than being generated. This framing has also been called \emph{next-utterance classification}.
Continuing this line of work are Hybrid Code Networks \cite{williams2017hybrid} which combine end-to-end learned models with domain-specific software to run parts of the business logic.
In a similar vein, dialogue transformers \cite{vlasov2019dialogue} have been proposed, which use self-attention over the dialogue history to make learned dialogue policies more robust to digressions. \cite{mosharrof2023zero} also learn a dialogue policy by training a seq2seq model which outputs the dialogue state, knowledge base queries and system actions at every turn to fulfill the goal of the user. The learned policy is shown to generalize well across unseen domains. 

The PolyResponse model \cite{henderson2019polyresponse} is a next-utterance classification approach that leverages transfer learning. The model is trained on a large corpus of unlabeled dialogue data and fine-tuned on domain-specific data to build a task-oriented system for a given domain.

However, it has been previously noted that collecting hand-annotated examples for training task-oriented dialogue systems is challenging \cite{budzianowski-etal-2018-multiwoz} and hence a lot of recent work aims at making the learning more data-efficient. \cite{pengsoloist} leverages transfer learning to make learning end-to-end dialogue models more data efficient, but where assistant responses are generated (as in a seq2seq model) rather than retrieved. 

\cite{jang2022gptcritic} leverage offline reinforcement learning (RL) to clone the behaviour exhibited by human agents in human-human conversations and use it to critique the actions taken by fine-tuned LLM-based agent at training time. On the other hand, \cite{hong2023zero} leverage task descriptions to generate relevant and diverse synthetic dialogues and use them to learn a dialogue agent via offline RL. This approach matches the performance of an LLM-based assistant using in-context learning, \cite{ouyang2022training} but with more concise and informative responses. However, this approach still requires fine-tuning an LLM which can be prohibitively time consuming for rapidly iterating on an implementation.

\subsection{Alternative Representations of Dialogue}
Other approaches have developed new representations and data structures to change how the task of building a dialogue system is formulated. 

 \cite{cheng-etal-2020-conversational} introduced a hierarchical graph structure  which represents the ontology of a task. Dialogue state tracking is then framed as a semantic parsing task over this structure.

\cite{andreasdataflow} represents the dialogue state as a dataflow graph, where the effect of each turn in a dialogue is to modify this graph.
They show that using this representation, a generic seq2seq model can match the performance of neural architectures specifically designed for dialogue state tracking. 
One similarity between our system and the dataflow approach is that both use the output of a model to generate instructions, and then deterministically execute some logic. 
However, there are significant differences in the two approaches. 
Our work also uses a graph representation of computational steps, but only to represent the business logic for a specific task, as designed explicitly be a developer. 
Additionally, the dataflow approach produces computational steps such as the \emph{refer} operation, which handles anaphora and entity resolution. 
In our system, the dialogue manager does not participate in language understanding, and coreference resolution is always handled implicitly, by including the conversation transcript in the LLM prompt and generating commands with the  arguments already fully resolved.
More generally, our approach uses the conversation transcript as a general-purpose representation of conversation state, while we use an explicit state representation only to track progress within the logic of a given task.

\subsection{Language Model ``Agents"}
 
Recently, researchers have explored using the in-context learning \cite{brown2020language} abilities of LLMs to have them act fully independently as dialogue systems \cite{yao2023react}, an approach sometimes referred to as LLM ``Agents".
This line of work assumes that the business logic required to complete a task is not known \emph{a priori} and must be inferred on-the-fly as a conversation progresses.
This approach explores the possibility of creating fully open-ended assistants which can help with an infinite number of tasks, with the caveat that the developer of the assistant does not control the task logic. 
Industrial dialogue systems, on the other hand, typically support a known set of tasks whose logic needs to be followed faithfully. 

Another approach is to frame the task of dialogue state tracking as a text-to-SQL problem \cite{hu2022context}. By leveraging in-context learning, this work demonstrates a dialogue agent that outperforms previously built fine-tuned agents \cite{shin-etal-2022-dialogue, lee-etal-2021-dialogue}. This approach leverages example conversations, which are retrieved at run time and inserted into an LLM prompt.  Finally, \cite{heck2023chatgpt} leverage in-context learning without any insertion of conversation examples in the prompt to explore the capability of ChatGPT as a task oriented dialogue agent. They find the system to be competitive to previously built zero-shot \cite{hu2022context} and few-shot \cite{lee-etal-2021-dialogue, shin-etal-2022-dialogue} dialogue agents.

\section{Architecture}
\label{sec:architecture}

Our architecture comprises three core elements: Business Logic, Dialogue Understanding, and Conversation Repair. 

When an end user sends a message to an assistant, the following takes place:

\begin{enumerate}
    \item{The dialogue understanding module interprets the conversation so far and translates the latest user message into a set of commands. }
    \item{The generated commands are validated and processed by the dialogue manager to update the conversation state.}
    \item{If the user message requires conversation repair, the corresponding repair patterns are added to the conversation state.}
    \item{The dialogue manager executes the relevant business logic deterministically, including any repair patterns, and continues executing actions until additional user input is required.}
\end{enumerate}

\subsection{Business Logic}
\label{sec:business_logic}

Business Logic describes the steps required to complete a specific task, such as transferring money.
Tasks are defined in a declarative format called \textbf{flows}.
A minimal flow comprises a description and a list of steps. 
The steps describe (i) what information is needed from the user (e.g. the amount of money and the recipient), (ii) what information is needed from APIs (e.g. the user’s account balance) and (iii) any branching logic based on the information that was collected. The following is an example definition of a minimal \texttt{transfer\_money} flow, which collects the recipient and amount from the user before initiating the transfer:

\begin{samepage}
{\small
\begin{verbatim}
transfer_money:
  description: send money to another
    account
  steps:
    - collect: recipient
    - collect: amount
    - action: initiate_transfer
\end{verbatim}
}
\end{samepage}
 
This task specification is created by the developer of the assistant. In addition to this core logic, they have to define the data types of the \texttt{recipient} and \texttt{amount} slots, and provide the templated utterances for the steps in the flow. 

\begin{samepage}
{\small
\begin{verbatim}
slots:
  recipient:
    type: text
  amount:
    type: float
responses:
  utter_ask_recipient:
    - text: Who are you sending money to?
  utter_ask_amount:
    - text: How much do you want to send?
\end{verbatim}
}
\end{samepage}

These two code snippets are all that is required to implement a task;
there is no training data required for language understanding.
The money transfer flow given here is a minimal example, but flows can include branching logic, function calls, calls to other flows, and more.
A more complex example can be found in appendix \ref{app:example_flow}.

Note that the flow definition does not make any reference to the user side of the conversation.
Neither dialogue acts nor commands are represented.
Business logic only describes the steps required to complete a task.
It does not specify \emph{how} the end user provides that information. 
While the flow only specifies the ``happy path", an assistant with this flow
can already handle a large number of conversations, including repair cases like
corrections, digressions, interruptions, and cancellations.
This is described in section \ref{sec:repair}.

\subsection{Dialogue Understanding}
\label{sec:dialogue_understanding}

In lieu of an NLU module, our system has a Dialogue Understanding module that leverages the in-context learning abilities of LLMs. Dialogue understanding, framed as a command generation problem, improves upon intent-based NLU in key ways:

\begin{itemize}
    \item While NLU interprets one message in isolation, DU considers the greater context: the whole running transcript of the conversation as well as the assistant’s business logic. Flow definitions and conversation state provide additional, valuable context for understanding users. This is especially useful for extracting slot values, which often requires coreference resolution.
    \item While NLU systems output intents and entities representing the semantics of a message, DU outputs a sequence of commands representing the pragmatics of how the user wants to progress the conversation \footnote{This is a similar idea to \cite{andreasdataflow}, where they also use a model to predict the \emph{perlocutionary force} of an utterance}.
    \item DU requires no additional annotated data beyond the specification of flows.
    \item While NLU systems assign a user message to one of a fixed list of intents, DU instead is generative, and produces a sequence of commands according to a domain-specific language and available business logic. This representation can express what users are asking with more nuance than a simple classification.
\end{itemize}

The following section illustrates these improvements with examples.

\subsubsection{Commands as a Domain-Specific Language}

The output of the Dialogue Understanding component is a short sequence of commands  (typically 1-3) describing how the end user wants to progress the conversation with the assistant. The allowed commands are shown in table \ref{tab:commands}.

\begin{table}
    \centering
    \begin{tabular}{l}
        \small \texttt{StartFlow(flow\_name)}\\
        \small \texttt{CancelFlow}\\
        \small \texttt{SetSlot(slot\_name, slot\_value)}\\
        \small \texttt{ChitChat}\\
        \small \texttt{KnowledgeAnswer}\\
        \small \texttt{HumanHandoff}\\
        \small \texttt{Clarify(flow\_name\_1, flow\_name\_2)}\\
    \end{tabular}
    \caption{A list of the commands which the dialogue understanding component can produce. Commands exist for starting and cancelling flows, for setting slots, for handing non-task dialogue (e.g. chitchat or content from a knowledge base), for handing over the conversation to a human agent, and for triggering an additional clarification step to handle disambiguation.}
    \label{tab:commands}
\end{table}

The following are some example user messages along with the corresponding command output.
In the simplest case, the user expresses a wish that directly corresponds to one of the defined flows:

\begin{userbox}
I want to transfer money
\end{userbox}

\begin{commandbox}
\small \texttt{StartFlow(transfer\_money)}
\end{commandbox}

Alternatively, the user may directly provide some of the required information.

\begin{userbox}
I want to transfer \$55 to John
\end{userbox}

\begin{commandbox}
\small \texttt{StartFlow(transfer\_money),\\ SetSlot(recipient, John), \\SetSlot(amount, 55)}
\end{commandbox}

Note that in the NLU-based approach, the values ``John" and 45 would typically be extracted as generic \texttt{person} and \texttt{number} entities, but in our system they are directly mapped to task-specific slots by the dialogue understanding component using the information from the flow definitions. 

Because the context of the conversation is taken into account, slots are filled correctly even in cases of pragmatic implicature, something that is very difficult to achieve with intent-based NLU:

\begin{botbox}
Are you traveling in economy class?
\end{botbox}

\begin{userbox}
sadly
\end{userbox}

\begin{commandbox}
\small \texttt{SetSlot(economy\_class, true)}
\end{commandbox}

Complex utterances can be represented faithfully as commands, when this would be difficult to achieve with an intent-based NLU approach.
For example, a user correcting a previous input and starting another task:

\begin{userbox}
Actually I meant \$45. Also what's my balance?
\end{userbox}

\begin{commandbox}
\small \texttt{SetSlot(amount, 45)}, \\ \texttt{StartFlow(check\_balance)}
\end{commandbox}

Section \ref{sec:repair} describes how corrections, interruptions, and digressions are subsequently handled by our system.

\subsection{Conversation Repair}
\label{sec:repair}

As described in section \ref{sec:business_logic}, a flow only specifies the steps required to complete a task.
It does not represent a graph of possible conversation paths.
Conversation repair defines a set of \emph{patterns}, which are meta flows that describe how the assistant behaves in conversations that deviate from the “happy path” of a flow.

We define the ``happy path" as any conversation in which the end user, every time they are prompted for information via a \texttt{collect} step, successfully provides that information, progressing to the next step in the business logic.

In production systems, end users frequently stray from the happy path, for example when they:
\begin{itemize}
    \item cancel the current interaction
    \item interrupt the current process to achieve something else before continuing
    \item request additional information
    \item insert an aside, e.g. ``just one moment"
    \item correct something they said earlier
    \item say something that requires further clarification
\end{itemize}

For these common situations, conversation repair provides patterns that are triggered through either specialized commands (e.g. \texttt{CancelFlow}) or when specific dialogue states are reached. For example, a specific pattern is triggered when a previously interrupted flow is resumed, because the interrupting flow finished or was cancelled. All patterns have a default implementation that can be overwritten by the developer of an assistant.

The following example requires a clarification step because the developer has created flows for multiple card-related tasks, and the user's opening message does not provide enough information to infer which one they want:

\begin{userbox}
card
\end{userbox}

\begin{commandbox}
\small \texttt{Clarify(freeze\_card, unfreeze\_card, cancel\_card)}
\end{commandbox}

\begin{botbox}
Would you like to freeze or unfreeze your card, or cancel it?
\end{botbox}

\begin{userbox}
cancel
\end{userbox}

\begin{commandbox}
\small \texttt{StartFlow(cancel\_card)}
\end{commandbox}

Implementing this behaviour using intent-based NLU is extremely challenging.
Words like `card' and `cancel' are not, on their own, indicative of a specific intent. 
Similarly, end users often start a conversation with very long messages that also require clarification.
These types of utterances tend to fit poorly into an intent classifier's taxonomy.
Furthermore, the dialogue manager would have to be programmed to handle many such sequences for various task combinations.
Our system handles disambiguation out of the box, without any additional effort from the developer.

\subsection{The Dialogue Stack}
\label{sec:dialoguestack}

Our system leverages a dialogue stack to process commands and execute business logic. 

The dialogue stack is a high-level representation of the conversation state. It maintains a Last-in-first-out (LIFO) stack of active flows, as well as the state of each individual flow.

The dialogue stack is also used to provide additional context to the dialogue understanding module, for example which slots the active flow needs to fill, along with their data types and allowed values. This helps the LLM generate the correct commands.

In turn, the commands generated by the LLM are used to manipulate the dialogue stack and conversation state with a specific set of operations. Commands can set slots and push new flows on to the dialogue stack. Beyond that, the commands the LLM generates do not directly manipulate the dialogue stack. They cannot directly remove or modify existing flows on the stack, eliminating the possibility of a malicious user overriding the business logic via prompt injection \cite{liu2023prompt}.

After all commands are processed, the dialogue manager takes the dialogue stack and executes the topmost flow deterministically. This execution pauses when it reaches a \texttt{listen} step. This way, executing business logic and complex operations on the dialogue stack alike are the domain of deterministic, developer-defined logic and not left to an LLM.

With this approach, our system leverages the powerful language understanding capabilities of LLMs while limiting their access to directly intervene in the business logic.

\subsection{Optional Components}

This section describes optional components which extend the core functionality of our system. 

\subsection{Contextual Rephrasing}
\label{sec:rephrasing}

The Contextual Response Rephraser is an optional component which can be used to improve the end user experience. 
It uses an LLM to rephrase the templated response to better account for the conversation's context, improving fluency. 
This is especially helpful for generic messages.
For example, responding to a request that is out of scope:

\begin{samepage}
\begin{userbox}
I'd like to add my partner to my credit card.
\end{userbox}

\begin{botsuggestionbox}
I’m sorry, I can’t help you with that.
\end{botsuggestionbox}

\begin{botbox}
Unfortunately, I cannot help you add users to your card.
\end{botbox}
\end{samepage}

While this can enhance the fluency and naturalness of conversations, there is a possibility that the rephrased response does not preserve the exact meaning of the templated utterance, depending on the choice of LLM, the prompt, and the sampling parameters. 
It is left to the developer of the assistant to decide if this is an acceptable trade-off.

\subsection{Flow Pre-Selection}
\label{sec:preselection}

As the number of tasks in an assistant increases, eventually the information required by the LLM exceeds the length of its context window. 
For these cases, the Dialogue Understanding component can be configured to pre-select a list of candidate flows to be included for in-context learning. 
This is achieved by using the latest user utterance to retrieve the $k$ most similar flows, as measured by embedding both the user utterance and the descriptions of the flows. 
This can potentially introduce a performance drop as the correct flow is not guaranteed to be among the top $k$.
We find that for sufficiently large $k$ (e.g. $k=20$), this error is negligibly small; in our experiments we were able to achieve a Hit@20 score of 100\%.

\subsection{Information Retrieval}
\label{sec:rag}

Industrial dialogue systems often combine task-oriented dialogue, built on business logic, with information retrieval.
While flows are ideal for multi-step tasks that rely on real-time data fetched from APIs, end users often have questions which can be answered based on static data. 
In our architecture, this is achieved by having the Dialogue Understanding component generate a \texttt{KnowledgeAnswer} command.
The `KnowledgeAnswer` command is handled as a pattern (a prebuilt meta flow, see section \ref{sec:repair}).
This pattern invokes an information retrieval component, which uses the latest user message as a query to a knowledge base, returning a selection of potentially relevant information. 
The results are then either presented directly to the user, or used as part of an LLM prompt to formulate an answer to the user. 
The latter approach is frequently called Retrieval-Augmented Generation \cite{gao2024retrievalaugmented}.
When the pattern triggered by a \texttt{KnowledgeAnswer} command has completed, it is removed from the top of the stack and the conversation proceeds as before.

\section{Evaluation}
\label{sec:evaluation}

Quantitatively evaluating an approach and system for building conversational AI is difficult \cite{bohus2009ravenclaw}.
Desirable qualities like ease of use and a quick learning curve can be studied by recruiting participants for a controlled study.
Adoption is another signal of success and can attest to the scalability of the solution.
In lieu of an absolute evaluation of the effectiveness of our system, we compare it in relative terms to an implementation in Rasa that follows an intent-based NLU approach. 
Our evaluation compares the effort required to achieve a similar level of functionality in both systems.

\subsection{Example Assistant}
As an example system we implemented a virtual assistant in English for a travel rewards bank account. The implementations, tests, and instructions for reproducing these experiments are available online\footnote{https://github.com/RasaHQ/tod-in-context-learning}. The assistant supports the following tasks:

\begin{table}[h]
    \centering
    \begin{tabular}{l}
        Transferring money\\
        Adding, listing, and removing known contacts\\
        Showing recent transactions\\
        Ordering a replacement card\\
        Searching for restaurants and hotels\\
        Setting up recurring payments\\
        Verifying an account\\
    \end{tabular}
    \label{tab:my_label}
\end{table}

\subsection{Metrics}
We evaluate both our system and intent-based implementations through a suite of 71 test conversations designed to test a variety of conversational abilities. 
Both implementations were built using a these tests as a guide, adopting a test-driven development approach.
The tests cover each of the tasks implemented in the assistant and a combination of happy paths and conversations involving repair.
The test conversations vary in length with a minimum of 2 turns, a maximum of 19, and an average of 7.8 turns.
Note that each test conversation represents a distinct conversation ``path", meaning that our tests are not designed to evaluate understanding of variations in phrasing, but rather variations in user behaviour.

The full set of test conversations is available together with the implementation in the github respository, and some example test conversations are shown in appendix \ref{apx:test_conversations}

As a proxy for ``effort”, we measure the lines of code and data in each implementation.
The implementation using the system described in this paper comprises 14 flows and 47 slots. 
The baseline implementation comprises 26 intents, 10 entity types, and 41 slots.

Table \ref{tab:ours_vs_nlu_total} compares the pass rate of both implementations on our test conversations as well as the lines of code and data in each implementation. 
For these experiments, we used GPT-4 as the LLM to power the dialogue understanding component, since it performed best in our experiments.
A comparison of the performance of different LLMs across languages and use cases is left to future work.

\begin{table}
    \centering
    \begin{tabular}{lrr}
        Implementation & ours & baseline \\ \hline
        Total test pass rate & 95.8\% & 47.9\%\\
        Lines of Code and Data & 1169 & 1713\\
    \end{tabular}
    \caption{Fraction of passing tests and number of lines of code and data for two implementations, one using the system described in this paper and a baseline system using intent-based NLU.}
    \label{tab:ours_vs_nlu_total}
\end{table}

\begin{table}
    \centering
    \begin{tabular}{lrrr}
      Category & \# tests & ours & baseline \\ \hline
      happy\_path & 24 & 24 & 22 \\
      cancellations & 6 & 6 & 2 \\
      corrections & 13 & 13 & 0 \\
      repetitions & 2 & 2 & 1 \\
      disambiguation & 4 & 3 & 0 \\
      input\_validation & 5 & 3 & 3 \\
      negations & 3 & 3 & 0 \\
      chitchat & 2 & 2 & 2 \\
      digressions & 9 & 9 & 2 \\
      knowledge & 3 & 3 & 2 \\
    \end{tabular}
    \caption{Number of passing tests in different categories, for ours and NLU-based implementations of similar effort.}
    \label{tab:ours_vs_nlu_by_category}
\end{table}

\section{Discussion}
\label{sec:discussion}

While informative, evaluating our system by comparing two implementations has a number of limitations and should not be overly relied upon as a quantitative guide. 
For one, a simple metric like ``lines of code and data" does not capture the complexity of the two implementations.
In our own subjective experience, the cognitive load of working with our system is much lower than with the intent-based NLU approach, and we hypothesise that this would be reflected in a controlled study involving external participants, manifesting for example as a reduced time to complete a given task. 
Second, our evaluation does not quantify the ability of either system to handle the lexical variation of real-world user input.
This would be better evaluated by deploying both implementations side-by-side in an A/B test. 

Nonetheless, we see distinctly that our system allows developers to build dialogue systems which can handle a multitude of conversation patterns with a modest amount of effort.
Note also that unlike the baseline, the implementation using our system \emph{only} addresses the happy paths, while corrections, digressions, and more are handled by out-of-the-box conversation repair. 
The disambiguation cases are worth noting as well, as these are handled automatically by our system, while presenting an extreme challenge for an intent-based NLU approach.
It is worth commenting on the impressive performance of our system on conversations involving corrections, especially in light of previous evidence that LLMs show poor performance on conversation repair \cite{balaraman2023thats}.
We believe this is because the repair-QA dataset, on which previous studies were based, is a far more challenging task that requires an LLM to produce free-form answers from the the world knowledge implicit in its parameters.
Handling corrections in our system only requires the LLM to reason over the conversation transcript and produce the appropriate command, with the correct slot value typically present verbatim within the prompt.

\section{Future Work}
\label{sec:future_work}

Important avenues for further work in this area include more comprehensive evaluation of the current system, including case studies of production systems.
In addition it would be valuable to study of the performance of different LLMs for the dialogue understanding task in various languages, including code-switching and multilingual applications. 
Also, given that the dialogue understanding task is well defined and requires an LLM only to produce a short sequence of known commands, it would be valuable to investigate whether smaller models can deliver similar performance at smaller cost and latency. 
Finally, it is crucial for production systems that these can be improved on the basis of interactions with real end-users\footnote{We use the term Conversation-Driven Development to describe this process\cite{Nichol2022}.}. 
There is a distinct difference in this regard between working with a system based on supervised learning (like intent-based NLU) versus one based on in-context learning.
Finding ways to (beyond prompt engineering) incorporate the signal from real user feedback is an active area of investigation. 

\section{Conclusion}
\label{sec:conclusion}

We have introduced and described a system for developing industrial task-oriented dialogue systems, combining the in-context learning abilities of LLMs with the deterministic execution of business logic.
We discussed the handling of both ``happy path" and conversation repair-type dialogues, and show how these can be handled effectively with relatively little effort, compared to the traditional approach of intent-based NLU.
We have made our system, as well as the example implementations, available for use and further study, and hope that this will facilitate the development of many successful conversational assistants.

\section*{Acknowledgments}
We would like to thank Oliver Lemon for providing feedback on a draft of this paper. We thank numerous colleagues at Rasa for support with implementing and testing the system. 

\bibliography{calm}
\bibliographystyle{acl_natbib}


\appendix

\onecolumn 

\section{Complex flow Example}
\label{app:example_flow}
{\small
\begin{verbatim}
transfer_money:
  description: This flow lets users send money to other accounts.
  name: transfer money
  steps:
    - collect: transfer_money_recipient
      description: the name of a person
    - id: "ask_amount" # we keep this id, because we use it for a jump
      collect: transfer_money_amount_of_money
      description: the amount of money without any currency designation
    - action: check_transfer_funds
      next:
        - if: not slots.transfer_money_has_sufficient_funds
          then:
            - action: utter_transfer_money_insufficient_funds
            - set_slots:
                - transfer_money_amount_of_money: null
                - transfer_money_has_sufficient_funds: null
              next: "ask_amount"
        - else: transfer_money_final_confirmation
    - id: transfer_money_final_confirmation
      collect: transfer_money_final_confirmation
      description: accepts True or False
      ask_before_filling: true
      next:
        - if: not slots.transfer_money_final_confirmation
          then:
            - action: utter_transfer_cancelled
              next: "END"
        - else: execute_transfer
    - id: execute_transfer
      action: execute_transfer
      next:
        - if: slots.transfer_money_transfer_successful
          then:
            - action: utter_transfer_complete
              next: "END"
        - else:
            - action: utter_transfer_failed
              next: "END"
\end{verbatim}
}
\section{Example Test Conversations}
\label{apx:test_conversations}

Here we show some example test conversations.

\subsection{Disambiguation}

\begin{userbox}
contact
\end{userbox}

\begin{botbox}
Would you like to add or remove a contact, or see a list of existing contacts?
\end{botbox}

\noindent
\begin{userbox}
add
\end{userbox}

\noindent
\begin{botbox}
What's the handle of the user you want to add?
\end{botbox}

\subsection{Correction}

\begin{userbox}
I want to transfer money
\end{userbox}

\begin{botbox}
Who do you want to transfer money to?
\end{botbox}

\begin{userbox}
John
\end{userbox}

\begin{botbox}
How much money do you want to transfer?
\end{botbox}

\begin{userbox}
\$100
\end{userbox}

\begin{botbox}
Please confirm: you want to transfer \$100 to John?
\end{botbox}

\begin{userbox}
Ah, scratch that, I meant \$110
\end{userbox}

\begin{botbox}
Ok, I have updated the amount to \$110
\end{botbox}

\begin{botbox}
Please confirm: you want to transfer \$110 to John?
\end{botbox}

\begin{userbox}
yes.
\end{userbox}

\begin{botbox}
Ok. Your transfer is complete.
\end{botbox}

\subsection{Context Switch}
\begin{userbox}
I want to send \$50 to Justin
\end{userbox}

\begin{botbox}
Just to confirm: you want to transfer \$50.00 to Justin?
\end{botbox}

\begin{userbox}
yes, and btw are there any fees associated with the sending?
\end{userbox}

\begin{botbox}
We will not charge you for this transfer
\end{botbox}

\begin{botbox}
Your transfer has been completed
\end{botbox}

\end{document}